\pdfoutput=1

\documentclass[11pt]{article}
\usepackage{graphicx}

\usepackage{ACL2023}
\usepackage{multirow}
\usepackage{times}
\usepackage{latexsym}
\usepackage{adjustbox}
\usepackage[T1]{fontenc}
\usepackage[utf8]{inputenc}

\usepackage{microtype}

\usepackage{inconsolata}
\DeclareUnicodeCharacter{0002}{ }
\title{"A Little is Enough": Few-Shot Quality Estimation based Corpus Filtering improves Machine Translation}

\author{Akshay Batheja, Pushpak Bhattacharyya\\[1ex]
CFILT, Indian Institute of Technology Bombay \\
\texttt{\{akshaybatheja, pb\}@cse.iitb.ac.in}
}

\begin{document}
\maketitle
\begin{abstract}
Quality Estimation (QE) is the task of evaluating the quality of a translation when reference translation is not available. The goal of QE aligns with the task of corpus filtering, where we assign the quality score to the sentence pairs present in the pseudo-parallel corpus. We propose a Quality Estimation based Filtering approach to extract high-quality parallel data from the pseudo-parallel corpus. To the best of our knowledge, this is a novel adaptation of the QE framework to extract quality parallel corpus from the pseudo-parallel corpus. By training with this filtered corpus, we observe an improvement in the Machine Translation (MT) system's performance by up to \textbf{1.8} BLEU points, for English-Marathi, Chinese-English, and Hindi-Bengali language pairs, over the baseline model. The baseline model is the one that is trained on the whole pseudo-parallel corpus. Our Few-shot QE model transfer learned from the English-Marathi QE model and fine-tuned on only 500 Hindi-Bengali training instances, shows an improvement of up to \textbf{0.6} BLEU points for Hindi-Bengali language pair, compared to the baseline model. This demonstrates the promise of transfer learning in the setting under discussion. QE systems typically require in the order of (7K-25K) of training data. Our Hindi-Bengali QE is trained on only 500 instances of training that is $1/40^{th}$ of the normal requirement and achieves comparable performance. All the scripts and datasets utilized in this study will be publicly available. 

\end{abstract}

\section{Introduction}
In recent times, Neural MT has shown excellent performance, having been trained on a large amount of parallel corpora~\cite{dabre}. However, not all language pairs have a substantial amount of parallel data. Hence, we have to rely on the noisy web-crawled corpora for low-resource languages. The task of \textbf{Parallel Corpus Filtering} aims to provide a scoring mechanism that helps extract good-quality parallel corpus from a noisy pseudo-parallel corpus. The task of \textbf{Quality Estimation} (QE) aims to provide a quality score for a translation when the reference translation is unavailable. We use Quality Estimation to assign the quality scores to the sentence pairs present in pseudo-parallel corpora and extract good-quality parallel sentences. We aim to improve the quality of Machine Translation for English(En)-Marathi(Mr), Hindi(Hi)-Bengali(Bn) and Chinese(Zh)-English(En) language pairs by using sentence-level QE-based corpus filtering. We observe that QE-based corpus filtering performs better than previously proposed methods.

Our contributions are:
\begin{enumerate}
    
    \item Adaptation of the QE framework, which is normally used for MT evaluation, to extract high-quality parallel corpus from pseudo-parallel corpus; to the best of our knowledge, this is a novel adaptation of the QE framework to extracting quality parallel corpus from the pseudo-parallel corpus.
    \item Demonstrating the promise of Few-Shot QE technique to generate training data for MT; a Hindi-Bengali QE model is trained with only 500 training instances transfer learned from an English-Marathi trained QE model; the filtered parallel data using this Hindi-Bengali QE system gives \textbf{0.6} BLEU point improvement over Hi-Bn MT system trained on the pseudo-parallel corpus.
    
    \item Demonstrating performance improvement of the Machine Translation systems by up to \textbf{1.8} BLEU points for English-Marathi, Hindi-Bengali and Chinese-English language pairs, over the model trained on the whole pseudo-parallel corpus.
    
\end{enumerate}

% Intro should contain the following items:
% \begin{itemize}
%     \item Problem statement
%     \item Motivation(this can be a separate section or can be a part of introduction)
%     \item Contribution
%     \item Roadmap
% \end{itemize}

\section{Related work}
\subsection{Parallel Corpus Filtering}
Neural Machine Translation (NMT) is extremely \textit{data hungry}~\cite{sutskever2014sequence, bahadanau, NIPS2017_3f5ee243}.
Recently, there has been a growing interest in the process of filtering noisy parallel corpora to enhance the data used for training machine translation systems. The Conference on Machine Translation (WMT) has organized annual Shared Tasks on Parallel Corpus Filtering (WMT 2018, WMT 2019, WMT 2020). ~\citet{lu-etal-2020-alibaba} proposed an approach that uses the Dual Bilingual GPT-2 model and the Dual Conditional CrossEntropy Model to evaluate the quality of the parallel corpus.~\citet{labse} proposed the LaBSE model, which is a multilingual sentence embedding model trained on 109 languages, including some Indic languages.~\citet{herold-etal-2022-detecting} mentioned different types of noise that can be injected in a parallel corpus and investigated whether state-of-the-art filtering models are capable of removing all the noise types proposed by~\citet{khayrallah-koehn-2018-impact}.\\ 
\indent Most recently,~\citet{akshay} used a combination of Phrase Pair Injection and LaBSE~\cite{labse} based Corpus Filtering to extract high-quality parallel data from a noisy parallel corpus. In contrast, we use QE-based filtering to extract high-quality data from noisy pseudo-parallel data. We observe that QE quality scores are superior to the LaBSE quality scores. 

\subsection{Quality Estimation}
Quality Estimation (QE) is the task of evaluating the quality of a translation when reference translation is not available. The state-of-the-art MonoTransQuest architecture, proposed by~\citet{ranasinghe-etal-2020-transquest}, builds upon XLM-R, a widely-used pretrained cross-lingual language model known for its ability to generalize to low-resource languages~\cite{conneau-etal-2020-unsupervised}.~\cite{kocyigit-etal-2022-better} proposed a combination of multitask training, data augmentation
and contrastive learning to achieve better and
more robust QE in a Parallel Corpus Mining setting. The Parallel Corpus Mining task aims to detect
the most similar texts in a large multilingual collection and perform sentence alignment. This motivates us to use QE in the Parallel Corpus Filtering task.

% Removed citation lines
% In the WMT 2020 shared task on Quality Estimation, MonoTransQuest achieved the highest sentence-level direct assessment score prediction~\cite{specia-etal-2020-findings-wmt}
%~\citet{lu-etal-2018-alibaba} performed a rule-based and word alignment-based quality scoring to remove noise from the parallel corpora.

% \section{Modeling}
% \subsection{Statistical Modeling}
% \subsection{Relating to deep learning}

% \section{Block diagram and architecture}
\begin{figure}[t]
% \begin{adjustbox}{width=\columnwidth}
\begin{center}
\includegraphics[scale=0.2]{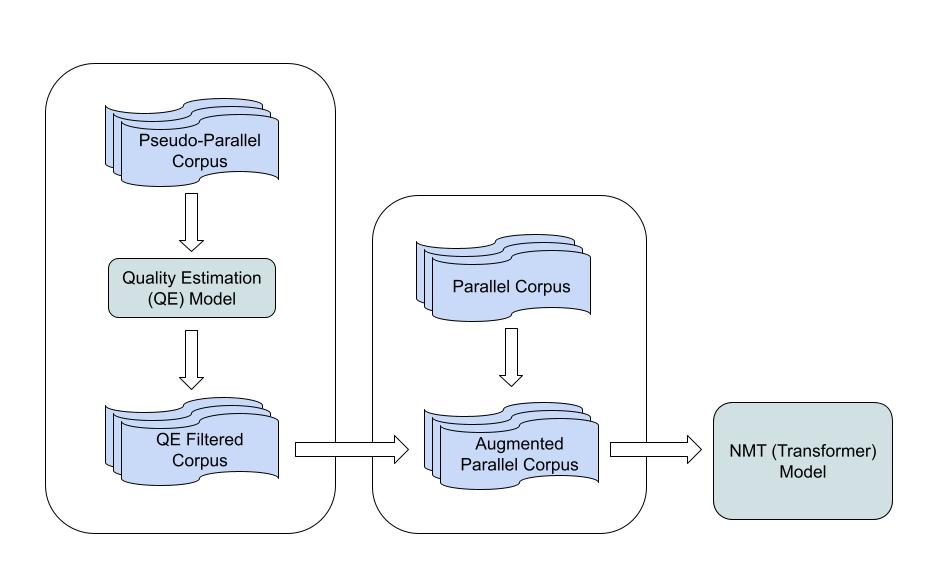}\caption{Quality Estimation based Filtering Pipeline}
\label{fig:1}
\end{center}
% \end{adjustbox}
\end{figure}
\section{Approaches}
We first discuss methods to extract good-quality parallel sentences from the pseudo-parallel corpus. Then we discuss a transfer learning-based filtering approach in few-shot settings.
% The pseudo-parallel corpus consists of Samanantar (English-Marathi) and WMT18 (Chinese-English) corpus. Samanantar is the largest publicly available corpora collection for Indic languages. This dataset contains sentence pairs of varying quality.
% \subsection{LaBSE based Fitlering}
\subsection{LaBSE based Filtering}
Language Agnostic BERT Sentence Embedding model~\cite{labse} is a multilingual embedding model that supports 109 languages, including some Indic languages. We generate the sentence embeddings for the source and target sides of the pseudo-parallel corpora using the LaBSE \footnote{\url{https://huggingface.co/sentence-transformers/LaBSE}} model.
Then, we compute the cosine similarity between the source and target sentence embeddings. After that, we extract good-quality parallel sentences based on a threshold value of the similarity scores.
\subsection{Phrase Pair Injection (PPI) with LaBSE-based Filtering}
~\citet{akshay} proposed a combination of Phrase Pair Injection~\cite{sen2021neural} and LaBSE-based Corpus Filtering to extract high-quality parallel data from a noisy parallel corpus. We train a PBSMT model on the noisy pseudo-parallel corpus using the \textit{Moses} \footnote{\url{http://www2.statmt.org/moses/?n=Development.GetStarted}} decoder. Then, we extract phrase pairs with the highest translation probability. Finally, we perform LaBSE-based filtering on these phrase pairs to remove poor-quality phrase pairs. We augment these high-quality phrase pairs with LaBSE-filtered parallel sentences.
\subsection{Quality Estimation based Filtering}
In this approach, we train the MonoTransQuest\footnote{\url{https://github.com/TharinduDR/TransQuest}}~\cite{ranasinghe-etal-2020-transquest} model and use it to generate the quality scores for the pseudo-parallel corpus of the corresponding language pair. Then, we extract high-quality parallel sentences from the pseudo-parallel corpus using a threshold quality score value.

\subsection{Few-shot Quality Estimation}
The Quality Estimation task requires human-annotated Direct Assessment scores for the corresponding language pairs. In few-shot settings, we fine-tune a pre-trained QE model for a high-resource language pair on QE data for the corresponding low-resource language pair to obtain a QE model for the low-resource language pair.
\section{Mathematical Preliminaries}
\label{app:Mathematical Prelimiaries}
\textbf{LaBSE scoring} \quad Let $D = \{(x_i, y_i)\}_{i=1}^{N}$ be a pseudo-parallel corpus with $N$ examples, where ${x_i}$ and ${y_i}$ represents $i^{th}$ source and target sentence respectively. We first feed all the source sentences present in the pseudo parallel corpus as input to the LaBSE\footnote{\url{https://huggingface.co/sentence-transformers/LaBSE}}~\cite{labse} model, which is a Dual encoder model with BERT-based encoding modules to obtain source sentence embeddings ($S_i$). The sentence embeddings are extracted as the l2 normalized \textbf{[CLS]} token representations from the last
transformer block. Then, we feed all the target sentences as input to the LaBSE model to obtain target sentence embeddings ($T_i$). We then compute cosine similarity $(score_i)$ between the source and the corresponding target sentence embeddings.
\begin{equation}
S_i=LaBSE\left(x_i\right) 
\end{equation}
\begin{equation}
T_i=LaBSE\left(y_i\right) 
\end{equation}
\begin{equation}
score_i=cosine\_similarity\left(S_i, T_i\right)    
\end{equation}
\textbf{QE scoring} \quad We feed ``$x_i [SEP] y_i$" as an input to the MonoTransQuest~\cite{ranasinghe-etal-2020-transquest} architecture which uses a single XLM-R model. The output of the [CLS] token is used as the input of a softmax layer that predicts the quality score $(score_i)$ of
the $i^{th}$ sentence pair $<x_i, y_i>$.
\begin{equation}
score_i=MonoTransQuest\left(x_i, y_i\right)    
\end{equation}

\section{Experimental Setup}
\label{experimental setup}
\begin{table}[htp]
\centering
\begin{adjustbox}{width=\columnwidth,center}
\begin{tabular}{l|l|r}
\hline
\textbf{Corpus Name} & \textbf{Language} &\textbf{Sentence}\\
&\textbf{Pairs} & \textbf{Pairs}\\
\hline
\multirow{3}{*}{\textbf{Parallel Corpus}} &Hindi-Bengali& 3.6M \\
&English-Marathi& 248K\\ 

&Chinese-English& 62K \\ \hline
% \multirow{2}{*}{\textbf{Pseudo Parallel Corpus}} &Hindi-Marathi&19.8L \\

% &English-Marathi&32.8L\\ \hline
\multirow{3}{*}{\textbf{Pseudo-Parallel Corpus}} &Hindi-Bengali&6.3M\\
 &English-Marathi&3.28M\\
 &Chinese-English&24.7M\\\hline
% Total & 25.9L \\ \hline
\end{tabular}
\end{adjustbox}
\caption{Dataset Statistics of  Parallel and Pseudo-Parallel Corpus for the task of Neural Machine Translation}
\label{lab:corpus}
\end{table}

\begin{table}[htp]
\centering
% \begin{adjustbox}{width=\columnwidth,center}
\begin{tabular}{l|r | r | r}
\hline
\textbf{Language} &\textbf{train} &
\textbf{dev} & \textbf{test}\\
\hline

English-Marathi& 26,000 & 1,000 & 1,000 \\ 
Chinese-English& 7,000 & 1,000 & 1,000\\
Hindi-Bengali& 440 & 50 & 10\\
\hline
% \multirow{2}{*}{\textbf{Pseudo Parallel Corpus}} &Hindi-Marathi&19.8L \\
% Total & 25.9L \\ \hline
\end{tabular}
% \end{adjustbox}
\caption{Dataset Statistics of human-annotated z-standardized Domain Adaptation (DA) scores for the task of Quality Estimation}
\label{lab:corpus2}
\end{table}

\begin{table*}[h]
% \small
\centering
% \begin{adjustbox}{width=\textwidth, center}
\begin{tabular}{lrrr}
\hline
\textbf{Technique} & \textbf{\# Sentence Pairs}& \textbf{En$\rightarrow$Mr} & \textbf{Mr$\rightarrow$En}\\ \hline
 \textbf{QE based Filtering} & 2.61M &9.4 & \textbf{17.7}\\
 \textbf{LaBSE + PPI-LaBSE  based Filtering} &  4.09M&\textbf{9.9} & 17.0\\
 \cite{akshay} & & &
 \\
 \textbf{LaBSE based Filtering} & 2.85M  &8.8 & 16.7 \\
 \textbf{Baseline} & 3.53M& 8.8 & 15.9\\ \hline
 % \textbf{Baseline} &  5.1 & 10.2 & 12.4 & 19.66 & 2.85   
\end{tabular}
    \caption{BLEU scores of En$\rightarrow$Mr and Mr$\rightarrow$En NMT models on FLORES101 test data. Here, we establish the efficacy of QE-based filtering in extracting a high-quality parallel corpus from En-Mr pseudo-parallel corpus. For actual instances of translations please refer to Appendix \ref{instances}.}
    \label{tab:2}
% \end{adjustbox}
\end{table*}

\begin{table*}[h]
% \small
\centering
% \begin{adjustbox}{width=\textwidth, center}
\begin{tabular}{lrr}
\hline
\textbf{Technique} & \textbf{\# Sentence Pairs}& \textbf{Zh$\rightarrow$En} \\ \hline
 \textbf{QE based Filtering} & 15.09M & 8.7\\
 \textbf{LaBSE + PPI-LaBSE  based Filtering} &  15.59M & 8.47\\ 
 \cite{akshay} & &\\
 \textbf{LaBSE based Filtering} &  15.57M& 8.29 \\
 \textbf{Baseline} &24.8M& 7.85\\ \hline
\end{tabular}
    \caption{BLEU scores of Zh$\rightarrow$En NMT models on FLORES101 test data. Here, we establish the efficacy of QE-based filtering in extracting a high-quality parallel corpus from Zh$\rightarrow$En pseudo-parallel corpus. For actual instances of translations please refer to Appendix \ref{instances}}
    \label{tab:2}
% \end{adjustbox}
\end{table*}
\begin{table*}[h]
% \small
\centering
% \begin{adjustbox}{width=\textwidth, center}
\begin{tabular}{lrrr}
\hline
\textbf{Technique} & \textbf{\# Sentence Pairs}& \textbf{Hi$\rightarrow$Bn} & \textbf{Bn$\rightarrow$Hi}\\ \hline

 \textbf{QE based Filtering} &7.77M & 13.28 & 21.06\\
 \textbf{LaBSE + PPI-LaBSE  based Filtering } &  8.73M& 13.24& 20.51\\
 \cite{akshay} & & &\\
 \textbf{LaBSE based Filtering} & 7.77M&
13.23 & 20.48\\
 \textbf{Baseline} & 10M& 12.91& 20.43\\ \hline
 % \textbf{Baseline} &  5.1 & 10.2 & 12.4 & 19.66 & 2.85   
\end{tabular}
    \caption{BLEU scores of Hi$\rightarrow$Bn and Bn$\rightarrow$En NMT models on FLORES101 test data. Here, we establish the efficacy of few-shot QE-based filtering using a pre-trained En-Mr model fine-tuned on Hi-Bn QE data to extract a high-quality parallel corpus from the Hi-Bn pseudo-parallel corpus. For actual instances of translations please refer to Appendix \ref{instances}}
    \label{tab:2}
% \end{adjustbox}
\end{table*}

\subsection{Dataset}
In all NMT experiments, we use two sets of corpus, namely, Parallel and Pseudo-Parallel corpus. The \textbf{Parallel corpus} consists of high-quality sentence pairs, while the \textbf{Pseudo-Parallel} corpus contains sentence pairs of varying quality. \\
\indent The En-Mr Parallel Corpus consists of the ILCI phase 1, Bible, PIB and PM-India corpus~\cite{ilci, bible, pmi}. The Zh-En Parallel corpus consists of ParaMed\footnote{\url{https://github.com/boxiangliu/ParaMed}} corpus. The Hi-Bn Parallel corpus is obtained from the OPUS\footnote{\url{https://opus.nlpl.eu/}} corpus repository. The En-Mr and Zh-En Pseudo-Parallel Corpus consist of the Samanantar~\cite{samanantar} and WMT18 Zh-En\footnote{\url{http://data.statmt.org/wmt18/translation-task/preprocessed/zh-en/}} corpus, respectively. The Hi-Bn Pseudo-Parallel Corpus consists of Samanantar and Tatoeba~\cite{tatoeba} corpus. The detailed data statistics are mentioned in table \ref{lab:corpus}.\\
\indent In QE experiments, we create a small corpus (500 instances) for Hindi-Bengali language pair that consists of human-annotated Domain Adaptation scores for each sentence pair annotated by three annotators. The pairwise Pearson Correlation between the three annotators of Hindi-Bengali QE is \textbf{0.68}, \textbf{0.61} and \textbf{0.67}. This indicates a good agreement between the three annotators. Please refer to \textbf{Appendix} \ref{annotation} for further annotation details. We use the QE data provided by~\citet{ranasinghe-etal-2020-transquest} and~\citet{deoghare-bhattacharyya:2022:WMT} for the Zh-En and En-Mr language pairs, respectively. The detailed 
QE data statistics are mentioned in table \ref{lab:corpus2}.\\
\indent For evaluation, we use the FLORES 101 test set which contains 1,012 sentence pairs for each language pair.
% \subsection{Baseline}
% We train the baseline models for English-Marathi, Hindi-Bengali, Chinese-English directly on their respective parallel corpus.

\subsection{Models}
\label{models}
We use MonoTransQuest model architecture to train the QE models. We use the Indic NLP library for preprocessing the Indic language data and  Moses for preprocessing the English language data. For Indic languages, we normalize and tokenize the data. For English, we lowercase and tokenize the data. We use a Transformer based architecture provided by OpenNMT-py library to train the NMT models for all our experiments.
The optimizer used was adam with betas (0.9, 0.98). The initial learning rate used was 5e-4 with the inverse square root learning rate scheduler. We use 8000 warmup updates. The dropout probability value used was 0.1 and the criterion used was label smoothed cross entropy with label smoothing of 0.1. We use a batch size of 4096 tokens. All the models were trained for 200,000 training steps. We use MonoTransquest\footnote{\url{https://github.com/TharinduDR/TransQuest}} model to train the sentence-level QE model. We start with a learning rate of 2e-5 and use 5\% of training data for warm-up. We use early patience over ten steps. We use a batch size of eight. The model architecture is mentioned in \textbf{Appendix} \ref{app:model arch}.\\
% \begin{itemize} respectively.
\noindent \textbf{Baseline} \quad We train the baseline NMT models on the whole pseudo-parallel corpus augmented with the parallel corpus for the corresponding language pairs.\\
\noindent \textbf{LaBSE based Filtering} \quad In this model, we use the LaBSE filtering with threshold 0.8 to extract good quality parallel sentences from the En-Mr, Hi-Bn and Zh-En pseudo-parallel corpus. Then we augment the parallel corpus with the LaBSE-filtered parallel sentences and train the respective NMT models. \\
\noindent \textbf{LaBSE + PPI-LaBSE based Filtering} \quad We extract LaBSE Filtered parallel sentences and phrases from the pseudo-parallel corpus and augment them with the parallel corpora to train the respective NMT models. \\
\noindent \textbf{Our Model, QE based Filtering} \quad We train the sentence-level QE model from scratch for En-Mr and Zh-En language pairs using their respective training data, Table \ref{lab:corpus2}. We use the English-Marathi pre-trained QE model for the Hi-Bn language pair and finetune it on Hi-Bn training data, Table \ref{lab:corpus2}. We compute quality scores for the noisy pseudo-parallel corpora using the trained QE models. Then, we extract high-quality sentence pairs from the pseudo-parallel corpus using the threshold values of -0.5, -0.4, and 0 for En-Mr, Zh-En, and Hi-Bn language pairs, respectively. We augment the extracted high-quality sentence pairs with the parallel corpus and train the respective NMT models.
% \subsection{LaBSE Filtering}

% \subsection{LaBSE + PPI-LaBSE Filtering}

% \subsection{QE Filtering}
\section{Results and Analysis}
We evaluate our NMT models using BLEU~\cite{papineni-etal-2002-bleu}. We use \textit{sacrebleu}~\cite{sacrebleu} python library to calculate the BLEU scores. Table \ref{tab:2} shows that QE based filtering model outperforms all other models for Hi-Bn, En-Mr and Zh-En language pairs.
The \textbf{QE based Filtering} model improves the MT system's performance by \textbf{0.85}, \textbf{0.6}, \textbf{1.8}, \textbf{0.37} and \textbf{0.63} BLEU points over the \textbf{baseline} model for Zh$\rightarrow$En, En$\rightarrow$Mr, Mr$\rightarrow$En,Hi$\rightarrow$Bn and Bn$\rightarrow$Hi, respectively. It also outperforms \textbf{LaBSE + PPI-LaBSE based Filtering} model by up to \textbf{0.7} BLEU points for Zh-En, En-Mr and Hi-Bn language pairs. The LaBSE + PPI-LaBSE based Filtering model performs better than QE based Filtering model for En$\rightarrow$Mr language direction. The LaBSE + PPI-LaBSE model, which is trained on nearly twice the amount of training data compared to the QE-based filtering model, can be a contributing factor to its better performance in En$\rightarrow$Mr.

The improvement in the performance of the Bn$\rightarrow$Hi QE-based filtered MT system is comparable to the En$\rightarrow$Mr and Zh$\rightarrow$En QE-based filtered MT model. The Hi-Bn QE model is trained with only 500 training instances transfer learned from En-Mr trained QE models. This demonstrates the promise of the few-shot QE technique to generate training data for MT.
% \section{Analysis}

\begin{table}[htp]
\centering
% \begin{adjustbox}{width=\columnwidth,center}
\begin{tabular}{l|r|r|r}
\hline
 \textbf{Technique} &\textbf{En-Mr} &\textbf{Hi-Bn} &\textbf{Zh-En}\\
 \hline
 
\textbf{LaBSE}& 0.44 & 0.51 & 0.2 \\
\textbf{QE}& \textbf{0.52} & \textbf{0.53} & \textbf{0.4}

\\
\hline
\end{tabular}
% \end{adjustbox}
\caption{Pearson Correlation between human annotated quality scores and quality scores computed using LaBSE and QE}
\label{lab:QE2}
\end{table}
\begin{table}[htp]
\centering
% \begin{adjustbox}{width=\columnwidth,center}
\begin{tabular}{r|r|r}
\hline
 \textbf{En-Mr} &\textbf{Hi-Bn} &\textbf{Zh-En}\\
 \hline
 
0.5& 0.37 & 0.28 \\
\hline
\end{tabular}
% \end{adjustbox}
\caption{Pearson Correlation between LaBSE and QE quality scores computed on the pseudo-parallel corpus for En-Mr, Hi-Bn and Zh-En 
language pairs respectively}
\label{lab:QE}
\end{table} 
We compute Pearson Correlation between human annotated quality scores and quality scores computed using LaBSE and QE, shown in Table \ref{lab:QE2}. The QE quality scores have a higher correlation with human annotated quality scores, compared to LaBSE quality scores for all 3 language pairs. Table \ref{lab:QE} shows the Pearson Correlation between LaBSE and QE quality scores for all 3 language pairs. We observe that the LaBSE quality score has a low correlation with the QE quality score and the QE quality score has a high correlation with the human annotated quality score. This establishes the superiority of QE over the LaBSE quality score.
% \subsection{Pearson's Correlation between LaBSE and QE quality-score}
% We compute Pearson Correlation between LaBSE and QE quality scores for En-Mr, Hi-Bn, and En-Zh language pairs.
% We observe that the coefficient value is positive but small for Hi-Bn and En-Zh language pairs. The MT models trained using QE-based filtering performs better than LaBSE-filtered MT models, table \ref{tab:2}, and both of the scoring mechanism are loosely correlated, table \ref{lab:QE}. We also compute Pearson Correlation between human annotated quality scores and quality scores computed using LaBSE and QE, table \ref{lab:QE2}. This indicates that QE quality scores are better than LaBSE quality scores and LaBSE quality scores require further investigation.

% \section{Case studies}
\section{Conclusion and Future Work}
We introduced a simple Quality Estimation based corpus filtering approach to extract high-quality parallel data from the noisy pseudo-parallel corpora. The takeaway from our work is that sentence-level QE-based filtering performs better than LaBSE-based filtering and helps improve the performance of NMT systems. We also show that few-shot QE models trained using a transfer learning-based approach can be used to extract good-quality parallel corpus from the pseudo-parallel corpus. Only $1/40^{th}$ of the normal data requirement (7K-25K) of QE training data achieves comparable performance for the Hindi-Bengali language pair. We also show that the QE quality score is superior to the LaBSE quality score.\\
\indent In the future, we plan to use the proposed corpus filtering technique for other language pairs. This will provide us with a general overview of how this filtering technique performs for multiple languages.
\section*{Acknowledgements}
We would like to thank the anonymous reviewers for their insightful feedback. We also express our gratitude towards Shivam Mhaskar, Sourabh Deoghare and other members of the Machine Translation group at CFILT, IIT Bombay, for their interesting and insightful comments.
\section*{Limitations}
Although our primary effort in this work was to extract as much parallel corpora as possible, the improvement in the performance has been found to be only marginal. The LaBSE and QE-based filtering experiments involve a hyper-parameter called "threshold quality score." To achieve optimal results, we conduct experiments with different values of this hyper-parameter. The proposed few-shot transfer learning technique requires a small amount of data that needs to be annotated by multiple annotators. 
% ACL 2023 requires all submissions to have a section titled ``Limitations'', for discussing the limitations of the paper as a complement to the discussion of strengths in the main text. This section should occur after the conclusion, but before the references. It will not count towards the page limit.
% The discussion of limitations is mandatory. Papers without a limitation section will be desk-rejected without review.

% While we are open to different types of limitations, just mentioning that a set of results have been shown for English only probably does not reflect what we expect. 
% Mentioning that the method works mostly for languages with limited morphology, like English, is a much better alternative.
% In addition, limitations such as low scalability to long text, the requirement of large GPU resources, or other things that inspire crucial further investigation are welcome.

\section*{Ethics Statement}
The aim of our work is to extract high-quality parallel corpus from a noisy pseudo-parallel corpus. The datasets that we used in this work are publicly available and we have cited the sources of all the datasets that we have used.  Publicly available datasets can contain biased sentences. We have also created a dataset for Hindi-Bengali few-shot QE. We briefly discuss the annotation guideline given to the annotators for the task in the \textbf{Appendix} \ref{annotation}.
% \section*{Acknowledgements}
% We would like to express our gratitude towards Shivam Mhaskar, Sourabh Deoghare, Jyotsana Khatri, and other members of the Machine Translation group at CFILT, IIT Bombay, for their interesting and insightful comments.

% 2. The BLEU score difference between QE and LABSE-PPI filtering is not significant for some language pairs. (Need to do significance test)

\bibliography{anthology,custom}
\bibliographystyle{acl_natbib}

\appendix

\section{Appendix}
\label{sec:appendix}

\subsection{Instances of Translations (Referred from Table \ref{tab:2})}
\label{instances}
The instances of translations for all 3 language pairs are provided in Table \ref{fig:i1}, \ref{fig:i2}, \ref{fig:i3}, \ref{fig:i4} and \ref{fig:i5}. 
\renewcommand{\figurename}{Table}
\renewcommand{\thefigure}{6}
\begin{figure*}[htp]
% \begin{adjustbox}{width=\columnwidth}
\begin{center}
\includegraphics[scale=0.75]{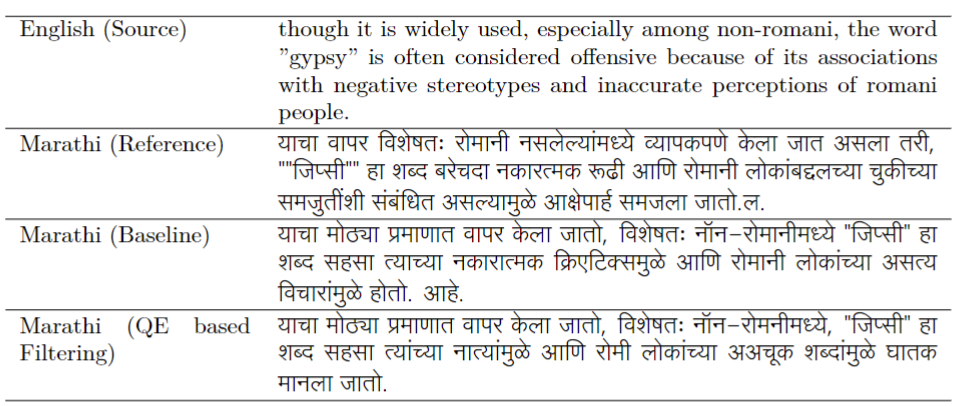}\caption{
Examples of NMT model output for En$\rightarrow$Mr}
\label{fig:i1}
\end{center}
% \end{adjustbox}
\end{figure*}

\renewcommand{\figurename}{Table}
\renewcommand{\thefigure}{7}
\begin{figure*}[htp]
% \begin{adjustbox}{width=\columnwidth}
\begin{center}
\includegraphics[scale=0.75]{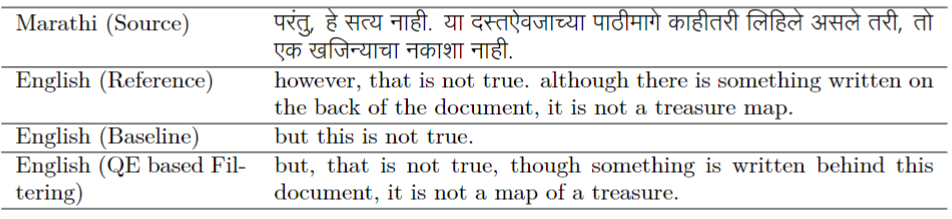}\caption{Examples of NMT model output for Mr$\rightarrow$En}
\label{fig:i2}
\end{center}
% \end{adjustbox}
\end{figure*}

\renewcommand{\figurename}{Table}
\renewcommand{\thefigure}{8}
\begin{figure*}[htp]
% \begin{adjustbox}{width=\columnwidth}
\begin{center}
\includegraphics[scale=0.75]{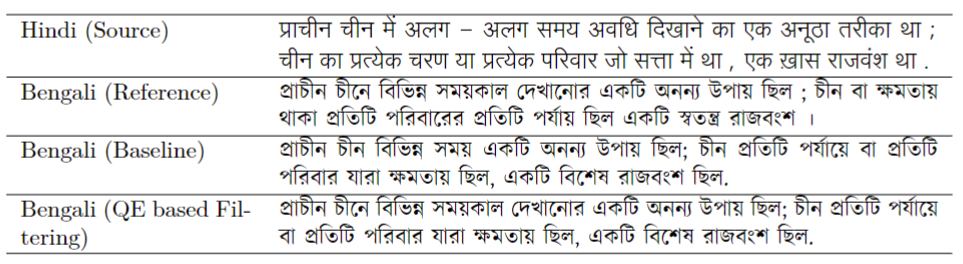}\caption{Examples of NMT model output for Hi$\rightarrow$Bn}
\label{fig:i3}
\end{center}
% \end{adjustbox}
\end{figure*}

\renewcommand{\figurename}{Table}
\renewcommand{\thefigure}{9}
\begin{figure*}[htp]
% \begin{adjustbox}{width=\columnwidth}
\begin{center}
\includegraphics[scale=0.75]{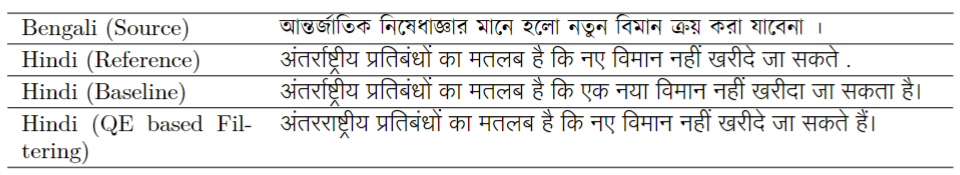}\caption{Examples of NMT model output for Bn$\rightarrow$Hi}
\label{fig:i4}
\end{center}
% \end{adjustbox}
\end{figure*}
\begin{figure*}[htp]
% \begin{adjustbox}{width=\columnwidth}

\renewcommand{\figurename}{Table}
\renewcommand{\thefigure}{10}
\begin{center}
\includegraphics[scale=0.75]{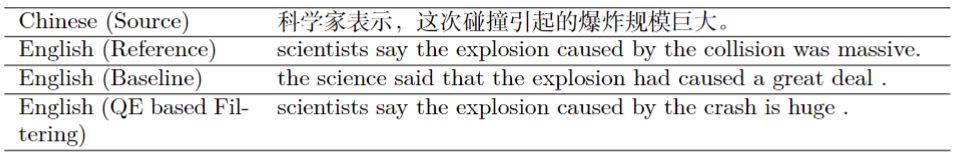}\caption{Examples of NMT model output for Zh$\rightarrow$En}
\label{fig:i5}
\end{center}
% \end{adjustbox}
\end{figure*}
\subsection{Model Architecture (Referred from section \ref{models})}
\label{app:model arch}
We use a Transformer based architecture to train English-Marathi, Hindi-Bengali and Chinese-English NMT models for all our experiments. The encoder of the Transformer consists of 6 encoder layers and 8 encoder attention heads. The encoder uses embeddings of dimension 512. The decoder of the Transformer also consists of 6 decoder layers and 8 decoder attention heads. We use MonoTransQuest architecture to train English-Marathi, Hindi-Bengali and Chinese-English QE models for all our experiments. We use a single Nvidia A100 GPU with 40 GB memory to train our NMT and QE models

% \subsection{Training Details (Referred from section \ref{models})}
% \label{app:training details}
% We used the Indic NLP library for preprocessing the Indic language data and  Moses for preprocessing the English language data. For Indic languages, we normalize and tokenize the data. For English, we lowercase and tokenize the data.
% We use the OpenNMT-py library to train the Transformer based NMT models. The hyperparameter values are selected using manual tuning. The optimizer used was adam with betas (0.9, 0.98). The initial learning rate used was 5e-4 with the inverse square root learning rate scheduler. We use 8000 warmup updates. The dropout probability value used was 0.1 and the criterion used was label smoothed cross entropy with label smoothing of 0.1. We use a batch size of 4096 tokens. All the models were trained for 200,000 training steps. We use MonoTransquest\footnote{\url{https://github.com/TharinduDR/TransQuest}} model to train the sentence-level QE model. We start with a learning rate of 2e-5 and use 5\% of training data for warm-up. We use early patience over ten steps. We use a batch size of eight. We use a single Nvidia A100 GPU with 40 GB memory to train our NMT and QE models.

\subsection{Annotation Details (Referred from section \ref{experimental setup})}
\label{annotation}
\subsubsection{Annotator Demographic}
For the Direct Assessment score annotation, we requested three native language speakers of Bengali who are well-versed in Hindi and have attended their graduate degrees in the Hindi language. They were aged between 25 to 42 and were paid for the time they spent on annotations.

\subsubsection{Guidelines}

The guidelines provided to the annotators for the Quality Estimation task are shown in Figure \ref{fig:i6}.
\renewcommand{\figurename}{Figure}
\renewcommand{\thefigure}{2}
\begin{figure*}[htp]
% \begin{adjustbox}{width=\columnwidth}

\begin{center}
\includegraphics[scale=0.9]{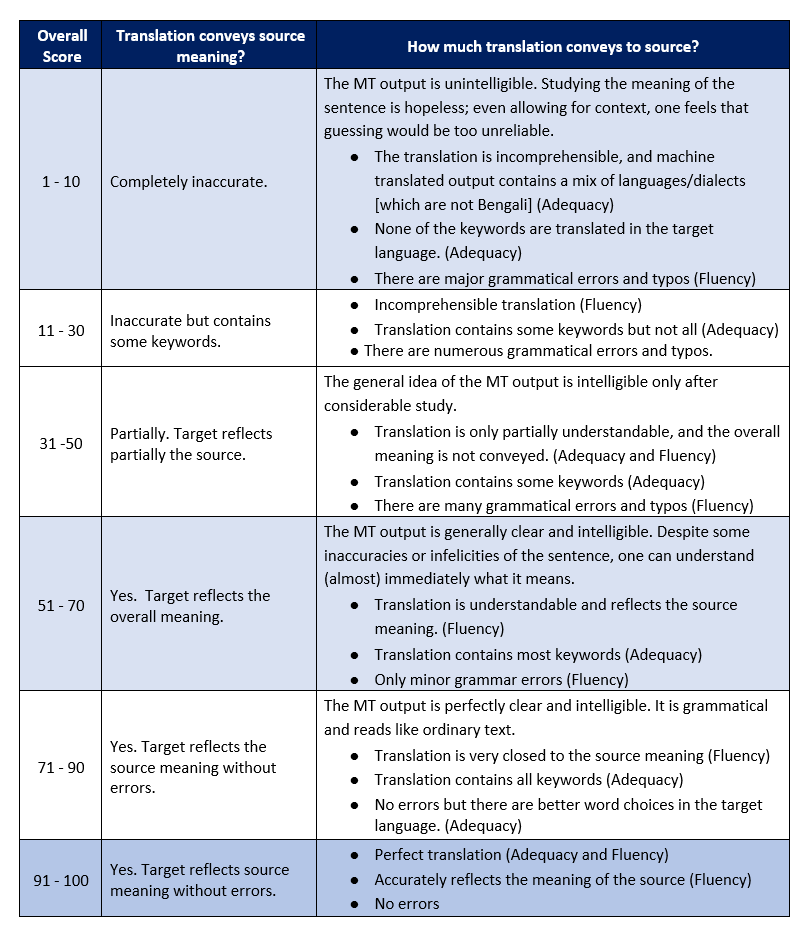}\caption{Evaluation Guidelines for the Hindi-Bengali Direct Assessment (for Quality Estimation) Task.  }
\label{fig:i6}
\end{center}
% \end{adjustbox}
\end{figure*}
\subsubsection{Dataset}
We create Hi-Bn QE data for our few-shot settings. We use 500 high-quality Hindi sentences from IIT Bombay English-Hindi parallel corpus~\cite{kunchukuttan-etal-2018-iit}. We use the Hindi-Bengali NMT model to generate translations for the 500 Hindi sentences. We provide this Hindi-Bengali parallel data to the annotators for the Direct Assessment Task. The Direct Assessment tasks require the annotators to score the MT translations as per the guidelines provided, Figure \ref{annotation}.
\end{document}